# An Empirical Comparison of SVM and Some Supervised Learning Algorithms for Vowel recognition

Rimah Amami, Dorra Ben Ayed, Noureddine Ellouze

*Department of Electrical Engineering, National School of Engineering of Tunis-ENIT*
*Rimah.amami@yahoo.fr, Dorra.mezghani@isi.rnu.tn, n.ellouze@enit.rnu.tn*

## Abstract

*In this article, we conduct a study on the performance of some supervised learning algorithms for vowel recognition. This study aims to compare the accuracy of each algorithm. Thus, we present an empirical comparison between five supervised learning classifiers and two combined classifiers: SVM, KNN, Naive Bayes, Quadratic Bayes Normal (QDC) and Nearst Mean. Those algorithms were tested for vowel recognition using TIMIT Corpus and Mel-frequency cepstral coefficients (MFCCs).*

## 1. Introduction

Learning Algorithms (or Learning Machine) [1] are a determinant elements for the accuracy of any recognition system. These algorithms are used usually, in order to recognize complex patterns and make efficient decisions. The Learning algorithms have set of applications like classifying, detecting, analysis, etc. Moreover, there are distinct types of Learning Algorithm i.e. the supervised Learning such as SVM, KNN, and Naive Bayes etc.

Learning algorithms has significant impact on the performance of Automatic Speech Recognition System (ASR), and so an effect on recognition rates. Therefore, the choice of a suitable learning algorithm play a key role in the success of any ASR system since they behave differently on different data sets.

In the other hand, one of the major developments in machine learning in the past decade is the combined methods. It consists of combining many moderately accurate component classifiers which may lead to finds a highly accurate classifier. The following fact opens the door for us to find combined classifiers which achieves a better generalization performance than individual classifiers.

In this context comes this article which presents a comparative study that shows the effect of using seven different supervised learning algorithms on the performance of our vowel recognition system. We evaluate the performance of SVMs, KNN, naive bayes, quadratic bayes normal, nearst mean and two hybrid methods on a multi-class recognition task.

The organization of this article is as follows. In Section 2, we present a brief overview of the different learning algorithms used. The experiments are described in section 3. We provide, in section 4, the results from a series of experiments to demonstrate the impact of the different algorithms for vowels recognition.

## 2. Methodology

Many different learning algorithms have been proposed and evaluated experimentally in various real-world applications domains, including speech recognition, handwritten character recognition, image classification and bioinformatics. In the following, we will review the different learning algorithm related to this study.

### 2.1. Support Vector Machine

The Support Vector Machine (SVM) method consists of constructing one or several hyperplanes in order to separate the different classes. Nevertheless, an optimal hyperplane must be found. Vapnik and Cortes [2] defined an optimal hyperplane as the linear decision function with maximal margin between the vectors of the two classes. The hyperplane can be described as:

$$W^T x + b = 0, x \in R^d$$

This margin is determinated by the training examples which are called support vectors (SV). Those SVs are the training examples close to the separation [3] and are the critical elements of the training set which are used to decide which hyperplane should be taken.

In the other side, for constructing a nonlinear SVM, Vapnik and Al. [4] suggest to apply the kernels tricks in purpose to maximum-margin hyperplanes. SVM can be used to learn polynomial classifiers, linear function, radial basic function (RBF) networks, and three-layer sigmoid neural networks by the following different kernel functions:

- Linear kernel function : $K(x_i, x_j) = x_i^T x_j$

- Polynomial kernel function : $K(x_i, x_j) = (\gamma x_i^T x_j + r)^d, \gamma > 0$

- RBF kernel function : $K(x_i, x_j) = \exp(-\gamma \|x_i - x_j\|^2), \gamma > 0$

- Sigmoid kernel function : $K(x_i, x_j) = \tanh(\gamma x_i^T x_j + r)$

This article chooses RBF as kernel function because it has the ability to map the examples into the wider space without complicating computation and because Gamma is the unique parameter to be set.

Moreover, the Platt's Sequential Minimal Optimization (SMO) method [5] was used, in this work, which allows a fast SVM training and with a fairly high number of samples. Besides, it solves the quadratic programming problem which occurs with SVM.

This study is based on multiclass phoneme recognition, thus, we use the "one-against-one" approach in which k (k-1)/2 classifiers are constructed and each one trains data from two different classes [6].

SVM is not an approximate algorithm; they are based on global optimization. Moreover, they deal with the overfitting problems which appear in a high dimensional space successfully. Thus, explain the robustness of SVM method and its tremendously appealing in various applications.

## 2.2. KNN

KNN, an acronym for K-Nearest Neighbor, is one of the simplest algorithms amongst the learning machine algorithms [7]. KNN is a non parametric supervised learning algorithm which means that it does not make any assumptions on the underlying data distribution [8]. This method has been used for classifying samples based on nearest training samples in the feature space. The main idea of KNN algorithm is to classify the samples based on the majority class of its nearest neighbors [9]:

$$Class = \arg\max_v \sum_{(X_i, y_i) \in D_z} I(v = y_i)$$

Where v is the class label, yi is the class label for the i[th] nearest neighbors and I is an indicator function that return the value of 1 if its argument is true and 0 otherwise.

Thus, the samples are assigned to the class of its K nearest neighbors [10][11]. Indeed, there are three key elements of the KNN approach: a set of labeled objects, a distance or similarity metric to compute distance between objects, and the value of k, the number of nearest neighbors. Choosing a suitable similarity function and an appropriate value for the parameter k is essential to make the recognition task successful.

KNN classification is an easy to understand and easy to implement classification technique. Despite its simplicity, it can perform well in many applications such as multi-modal classes as well as applications in which an object can have many class labels.

## 2.3. Naive Bayes

Despite of its simplicity, Naive Bayes (NB) can often outperform more sophisticated classification methods. The Naive Bayes is a classification algorithm based on bayes rule applied to categorical data. This supervised classifier aims to reduce the intractable sample complexity by making a conditional « naive » assumption of class in order to reduce computation in evaluating [12]:

$$P(X \mid C_i) P(C_i)$$

Indeed, this classifier simply computes the conditional probabilities of the different classes, given the values of attributes and then selects the class with the highest conditional probability [13]. It means that a sample is classified into a class label Ci if and only if it is the class that maximizes [14]:

$$f(X \mid C_i) P(C_i)$$

Where $f(X \mid C_i)$ is the conditional distribution of X for class $C_i$ objects, and $P(C_i)$ is the probability that an object will belong to class $C_i$.

The naive Bayes algorithm is hugely appealing because of its simplicity and robustness. It is one of the oldest formal classification algorithms, which is easy to construct and don't need any complicated iterative parameter estimation schemes. It is widely used in areas such as pattern recognition classification.

## 2.4. Quadratic Bayes Normal

The Quadratic Bayes Normal classifier (QDC) is also a common supervised learning algorithm for classification [15] [16]. It's a Bayes-Normal-2 algorithm which aims to separate measurements of two or more classes of samples by a quadric surface. This is the bayes rule which estimate for each class a normal distribution, with a separate covariance matrix yielding quadratic decision boundaries [17] [18]. This was done by estimating covariance matrix C for the scatter matrix S by:

$$C = (1 - \alpha - \beta)S + \alpha \, diag(S) + \frac{\beta}{n} \sum diag(S)$$

In which n is the dimensionality of the feature space and $\alpha = \beta = 10^6$.

## 2.5. Nearest Mean

Under Nearest Mean algorithm, the samples are assigned to the class of the nearest mean. A sigmoid function over the distance is used to estimate the posterior probabilites [17]. In the nearest mean classifier, the Euclidean distance from each class mean is computed for the decision of the class of the test sample [19]:

$$\|x - \mu_i\|^2 = (x - \mu_i)^T (x - \mu_i)$$

Where x is the sample and $\mu_i$ is the ith reference vector, see figure 1.

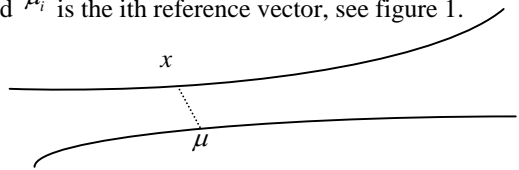

**Figure 1.** Illustration of the Euclidean distance between an observation x and a reference vector $\mu$

After computing the distance to each class mean, the test sample is classified into the class for which the Euclidean distance between the sample and the class mean is minimum [20].

## 3. Experiments

As the *No Free Lunch Theorem* suggests, there is no universally best learning algorithm. Even the best algorithm (i.e. decision tree, SVM, ect.) may have low performance on some problems, while a algorihtm that has been known to have poor average performance may performs well on a few problems. This paper presents results of an empirical comparison of seven supervised learning algorithms for vowel recognition. We evaluate the performance of SVM, KNN, Naive Bayes, QDC, Nearst Mean, KNN-QDC, and KNN-QDC- Naive Bayes (see section 2) on TIMIT vowel data for a multiclass recognition problem. Then we compare the outcome of each classifier to try improves their performance by combining algorithms which have the higher error rates. We find that, indeed, combining classifiers gives improvement in accuracy.

### 3.1. The architecture of the recognition system

The aim of this research is to obtain accurate recognition of vowels from acoustic information using different supervised learning algorithms. Afterwards, we compare those results in order to find the most suitable and accurate learning algorithm for our data sets in future researches. Thus, our vowel recognition system go through different stage which consists of : (1) Conversion from spectrogram to MFCC spectrum using MFCC analysis where for each 16ms frame, the feature extractor compute a vector of MFCC features  (2) apply a learning algorithm to the MFCC vector (3) get the vowel recognition rates using a suitable decision strategy, see figure 2.

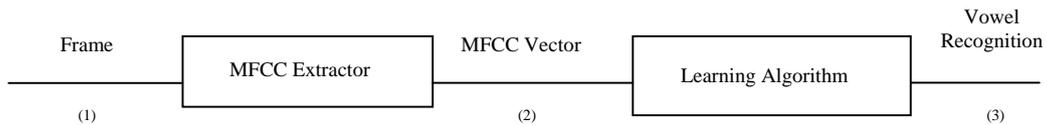

**Figure 2.** The vowel recognition system

### 3.2. Mel-Frequency Cepstral Coefficients

Mel-frequency cepstral coefficients (MFCC) is the most often used in speech recognition filed because of its robustness [21][22]. The main idea of this algorithm consider that the MFCC are the cepstral coefficients calculated from the mel-frequency warped Fourier transform representation of the log magnitude spectrum.
The Delta and the Delta-Delta cepstral coefficients are an estimate of the time derivative of the MFCCs. In order to improve the performance of speech recognition system, an improved representation of speech spectrum can be obtained by extending the analysis to include the temporal cepstral derivative; both first (delta) and second (delta-delta) derivatives are applied. Those coefficients have shown a determinant capability to capture the transitional characteristics of the speech signal that can contribute to ameliorate the recognition task.
In this study, each vowel is characterized directly by a set of three vectors which specify the three middle windows of the MFCC representation [23]. Each window of each vowel phoneme is represented by the 12 MFCC coefficients, the 12 Delta and the 12 Delta-Delta

### 3.3. Data Set

For the experiments, we used a set of supervised learning algorithms taken from the implementations available in Matlab Toolbox PrTools [24] with their default parameter values and for SVM we used LibSVM toolbox [25]. The performance of vowel recognition system is verified by experiments with a TIMIT speech corpus licensed from the Linguistic Data Consortium at the University of Pennsylvania. Phonemes are extracted from all sentences of the first dialect DR1. For the recognition task, we select twenty vowel phonemes: */ay/, /ae/, /ow/, /ey/, /ao/, /uw/, /aa/, /ah/, /aw/, /ax/, /ax-h/, /axr/, /eh/, /er/, /ih/, /ix/, /iy/, /oy/, /uh/, and /ux/.*

### 3.4. Combined Classifiers

Can we do better than individual classifiers if the classifiers were combined? Much work has been done in the past decade about the classifiers fusion methods which can be an efficient way to boost the performance generalization of machine learning in order to obtain improved recognition results [26][27].

Indeed, combining classifiers aims to solve accuracy systems problems by the combination of two (or more) different classifiers. Thus, the combined classifiers exploit the advantage of the strengths of different classifiers by applying them to problems they can efficiently solve.

However, each classifier is operating by its own properties, which presents a problem in decision combination. Hence, it should be pointed out that a set of classifiers such as KNN, QDC, NaiveBayes can be combined by several rules like Majority Voting, Maximum, Product, Mean etc., and the objective is to get a new classifier in order to improve the overall accuracy. Here we are interested in the majority voting strategy which is considered as one of the most important fusion strategy in this research area.

Recently, Lam and Suen [28] [29] have studied the accuracy of the majority vote strategy where they state that the majority vote is guaranteed to do better than an individual classifier when the classifiers have an accuracy greater than 50%. They find, also, that in speech recognition applications, the majority voting strategy is the most appropriate strategy to be used.

In a recognition task, the majority voting strategy consists of counting the votes for each class over the input classifiers and selecting the majority class.

The choice of classifiers to be combined was based on the accuracy of each individual classifier. Thus the classifiers where the error rates were greater than 50% will be combined with each other. In the first place, only two classifiers will be combined then in the second place we will combine three classifiers in order to see the impact of the number of classifiers combined on the recognition system.

### 4. Results and Discussion

In this section we provide experimental results to show how each individual and combined classifier work on a multiclass vowel recognition task. Based on table 1, SVM classifier had achieved the best performance amongst the different classifiers evaluated. However, QDC was the classifier which had achieved the lowest accuracy rates in all experiments. And, hence comes the idea to combined QDC with others classifiers in order to observe if can we can do better? Indeed, we notice a 16% improvement over this classifier. The error rates of the remaining classifiers (KNN, Nearest Mean, and Naïve Bays) were greater than 50%.

Meanwhile, some classifiers must have parameters to be settled such as KNN. Therefore, we were, first, attempted to investigate the best choice for the KNN classifier on a case study of vowel dataset.

Furthermore, the choice of K is primordial in building the KNN model. K can strongly influence the recognition rates. For any given problem, a small value of k will lead to a large variance in predictions. Alternatively, larger values of k reduce the effect of noise on the classification, but make boundaries between classes less distinct. Thus, k should be set to a value large enough to minimize the probability of misclassification and small so that the K nearest points are close enough to the query point. The optimal value for k can be regarded as a value which achieves the right tradeoff between the bias and the variance of the model. Generally, K must be a positive integer and typically small.

The figure 3 presents the recognition rates with different value of K in order to find the suitable choice for our vowels database.

An Empirical Comparison of SVM and Some Supervised Learning Algorithms for Vowel recognition
Rimah Amami, Dorra Ben Ayed, Noureddine Ellouze

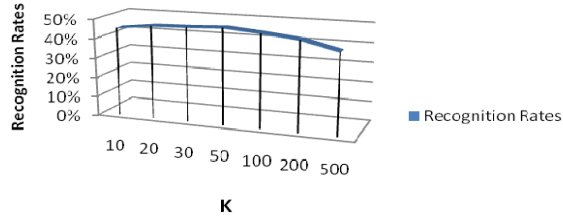

**Figure 3.** KNN recognition Rates for different K value

As seen in the figure 3, 50-NN gives the best recognition rates. We notice also that with a large value of K, the recognition rates decrease. Thus, in the next experiments, we choose K=50.

In the following experiments, the SVM classifier is compared with 6 classifiers in order to evaluate the performance of the different supervised algorithms on vowel data, see table 1.

Otherwise, for the nonlinear SVM approach, we choose the RBF Kernel trick. This choice was made after a previous study done on our datasets with different kernel tricks (Linear, Polynomial, and Sigmoid). As the classification performance of SVMs is mainly affected by its model parameters particularly the Gaussian width Gamma and the regularization parameter C, we set, for all experiments, gamma as a value within 1/k where k is the number of features ($\gamma = 0.027$) and C as value within 10.

**Table 1.** % Test error of different supervised learning Classifiers

| Classifiers | Error Rates |
|---|---|
| 50-KNN | 51 |
| QDC | 64 |
| Naïve Bayes | 57 |
| KNN-QDC | 61 |
| Nearest Mean | 60 |
| KNN-QDC-Naïve Bayes | 54 |
| SVM | 48 |

Based on table 1, the SVM algorithm performs slightly better than K-NN algorithm on our vowel data with 48% test error. We observe, also, that the SVM algorithm performs about 10% better than Naive Bayes, and about 16% better than QDC. Meanwhile the table 1 shows that QDC classifier achieves the highest test error within 64%.

Hence, in order to improve the performance of QDC algorithm, we combined it with several classifiers used in this study. Indeed, as is shown in table 1, the combined KNN-QDC leads to decrease slightly the error rates. Then, when the number of combined classifier was increased by 3 we observed, also, that the combined KNN-QDC-NB has obviously decreased the error rate about 10%. It should be pointed that even when we combined classifiers with accuracy less than 50%, their accuracy is higher than the individual classifiers (Naive Bayes, QDC). These error rates improvements can be considered as a foregone fact since the main role of combined classifier is to improve the system's accuracy.

In the other hand, we observe that the Nearest Mean and the QDC classifiers perform hard comparing to the individual classifiers used by giving the highest error rates in all experiments.

Overall, we are comfortable to say that the performance of SVM is better than the different supervised algorithms used on this study. We would like to emphasize that the purpose of our empirical experiments is not to argue that SMV is the best supervised algorithm but rather to illustrate that SVM gives a better accuracy on our vowel data.

## 5. Conclusion

In this article, we proposed an empirical evaluation of the performance of some learning algorithms for a multiclass vowel recognition task. The experimental results reveal that SVM-RBF kernel provided the best performance learning accuracy compared to NB, KNN, QDC and even the combined classifiers used in this study. Although some methods clearly perform better or worse than other methods, their performance can significantly improves in further works by using different feature representations such as PLP and Rasta-PLP and some famous combined classifiers such as the Boosting classifiers.

## 6. Outlook

We conduct this study in aim to look for the suitable model for our phoneme recognition task. This comparative study of several learning algorithms has revealed that SVM was the best supervised learning algorithm to deal with our problem. This was an expected outcome and it was just confirmed with results reported in this study. SVM is known to achieve excellent performance that would have been difficult to obtain with others classifiers. However, it has been revealed more interesting fact: the combining classifier to improve accuracy. In the future, we plan to extend and refine our model using different features representations and combining SVM with Boosting algorithms to investigate the performance improvements.